\documentclass[letterpaper, 10 pt, conference]{ieeeconf}  

\IEEEoverridecommandlockouts                              

\usepackage{graphics} 
\usepackage{epsfig}
\usepackage{mathptmx} 
\usepackage{times} 

\usepackage[ruled,vlined,linesnumbered]{algorithm2e}
\usepackage{amsmath,amssymb,amsfonts}
\usepackage{subcaption}
\usepackage{algorithmic}

\newcommand{\tf}{{t_{\rm f}}}

\title{\LARGE \bf
Accelerating Kinodynamic RRT* Through Dimensionality Reduction
}

\author{Dongliang Zheng$^{1}$ and Panagiotis Tsiotras$^{2}$
\thanks{This work has been supported by NSF awards IIS-1617630 and  IIS-2008695}
\thanks{$^{1}$Dongliang Zheng is with School of Aerospace Engineering,
        Georgia Institute of Technology, Atlanta, GA 30332, USA
        {\tt\small dzheng@gatech.edu}}%
\thanks{$^{2}$Panagiotis Tsiotras is with School of Aerospace Engineering and Institute for Robotics and Intelligent Machines, Georgia Institute of Technology, Atlanta, GA 30332, USA
        {\tt\small tsiotras@gatech.edu}}%
}

\begin{document}

\maketitle
\thispagestyle{empty}
\pagestyle{empty}

\begin{abstract}

Sampling-based motion planning algorithms such as RRT* are well-known for their ability to quickly find an initial solution and then converge to the optimal solution asymptotically. 
However, the convergence rate can be slow for high-dimensional planning problems, particularly for dynamical systems where the sampling space is not just the configuration space but the full state space.
In this paper, we introduce the idea of using a partial-final-state-free (PFF) optimal controller in kinodynamic RRT* \cite{Dustin2013Kinodynamic} to reduce the dimensionality of the sampling space.
Instead of sampling the full state space, the proposed accelerated kinodynamic RRT*, called Kino-RRT*, only samples part of the state space, while the rest of the states are selected by the PFF optimal controller. 
We also propose a delayed and intermittent update of the optimal arrival time of all the edges in the RRT* tree to decrease the computation complexity of the algorithm. 
We tested the proposed algorithm using 4-D and 10-D state-space linear systems and showed that Kino-RRT* converges much faster than the kinodynamic RRT* algorithm.

\end{abstract}

\section{Introduction}

Robotic motion planning with the goal of finding a dynamically feasible and optimal trajectory for the robot through an environment with obstacles has gained much progress over the past decades.
As a fundamental problem in robotics applications, it is still a challenging problem to solve when the environment is complex with irregular obstacles and the dynamics of the robot are to be considered \cite{lavalle2011motion}.

Sampling-based motion planning algorithms, such as rapidly exploring randomized trees (RRTs) \cite{lavalle2001randomized}, have been developed to solve planning problems in high-dimensional continuous state spaces by incrementally building a tree through the search space.
The asymptotic optimal variant of RRT, namely RRT*~\cite{Karaman2011Sampling}, almost surely converges asymptotically to the optimal solution.
RRT* is well-suited for planning in high-dimensional spaces and obstacle-rich environments.
Many applications of RRT* have been studied in recent years~\cite{Karaman2011Anytime, gonzalez2015review, gammell4asymptotically}.

One limitation of RRT* is that it requires any two points sampled in the planning space to be connected with an optimal trajectory. Thus, many works on RRT* consider robots with simple dynamics~\cite{Karaman2011Anytime, Karaman2010Optimal} or assume a holonomic model and connect sampled points with straight lines~\cite{Gammell2014InformedRRT}.
For robots with differential constraints, the optimal trajectory between two states is obtained by solving a two-point boundary value problem (TPBVP), which is a non-trivial undertaking for complex nonlinear systems. 
The solution to this local TPBVP is also referred to as the steering function. 
A version of the RRT* algorithm that explicitly considers differential dynamics is the kinodynamic RRT*~\cite{Dustin2013Kinodynamic, Karaman2010Optimal}.

Solving TPBVPs is the computationally dominant component of kinodynamic RRT*, and thus researchers have looked into more efficient ways to solve these TPBVPs.
A steering function based on LQR is used in~\cite{Perez2012LQR}.
A fixed-final-state free-final-time controller that optimally connects any pair of states is used in~\cite{Dustin2013Kinodynamic}. 
Learning-based RRT* algorithms are introduced in~\cite{Wolfslag2018RRT-CoLearn,zheng2021sampling,Chiang2019RLRRT}, where the TPBVP is solved using supervised learning~\cite{Wolfslag2018RRT-CoLearn,zheng2021sampling} and reinforcement learning~\cite{Chiang2019RLRRT}.

Another challenge of RRT* is the slow convergence rate of the solution to the optimal one, which is especially evident for the kinodynamic case where the sampling space is not just the configuration space but the full state space.
Heuristic and informed sampling methods have been developed to improve the convergence rate~\cite{Gammell2014InformedRRT, arslan2015machine, janson2015fast}. 
However, these methods only consider the geometric planning problem and the dynamics of the robot is not considered.
Good heuristics for improving the convergence of kinodynamic RRT* is still an open research problem~\cite{paden2017verification,yi2018generalizing}.

\begin{figure}[htb]
    \centering
    \begin{subfigure}[b]{0.49\columnwidth}
         \centering
         \includegraphics[width=1\columnwidth]{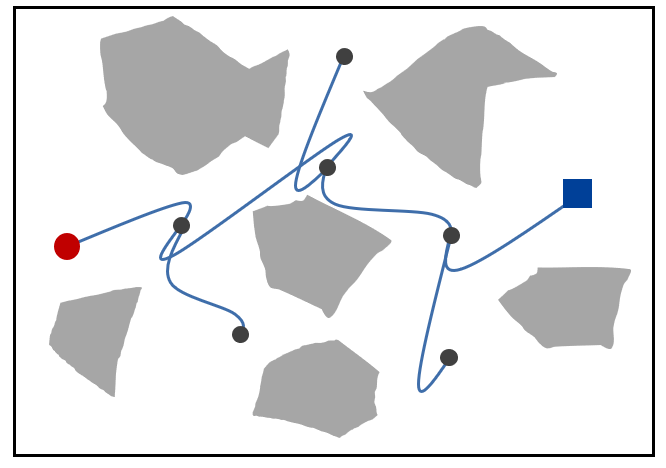}
         \caption{}
     \end{subfigure}
     \begin{subfigure}[b]{0.49\columnwidth}
         \centering
         \includegraphics[width=1\columnwidth]{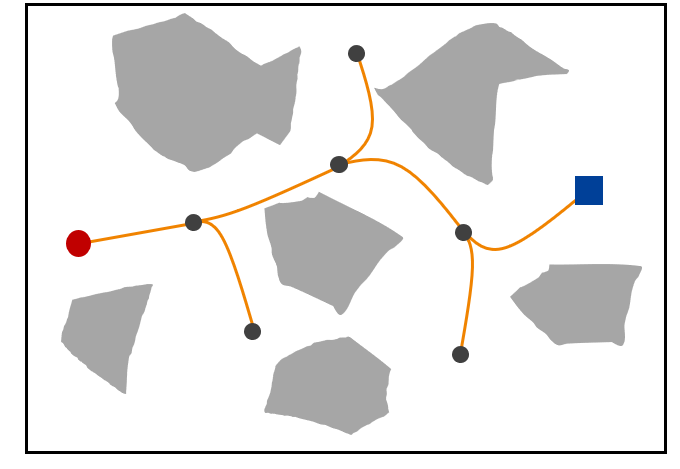}
         \caption{}
     \end{subfigure}
        \caption{Motivation of the partial-final-state-free (PFF) optimal controller. (a) Existing kinodynamic RRT* algorithms sample the full state space, which results in inefficient trajectories. (b) Kino-RRT* with a PFF controller samples the reduced state-space to improve convergence performance. }
        \label{IllustrationOFpff}
\end{figure}

In this paper, we build on previous work on the kinodynamic RRT*~\cite{Dustin2013Kinodynamic} and propose a new algorithm, called Kino-RRT*, which shows faster convergence.
We propose the idea of using a partial-final-state-free (PFF) optimal controller to reduce the sampling dimension of the state space.
The motivation is illustrated in Figure~\ref{IllustrationOFpff}. 
Instead of randomly sampling the full state space, the proposed Kino-RRT* only samples part of the state space.
The rest of the states are selected by the PFF optimal controller. 
Because part of the final state is computed by the PFF optimal controller to optimize the cost function, Kino-RRT* samples in the state space with reduced dimension.
The method can also be interpreted as a heuristic for state-space sampling.
Choosing the partial-free final states by the PFF optimal controller is more efficient than uniformly random sampling, and thus the resulting algorithm achieves faster convergence. 
We derive an analytical solution of the PFF optimal controller for the case of linear systems.
Note, however, that the idea of using PFF controller in kinodynamic RRT* is not limited to linear systems. 
It can be adopted similarly to~\cite{Xie2015Towards, Wolfslag2018RRT-CoLearn, zheng2021sampling} to deal with nonlinear dynamics as well.

Finding the optimal arrival time for the TPBVP in the kinodynamic RRT* requires solving a root-finding problem of a high-order polynomial.
Because the TPBVP is required to be solved repeatedly, the root-finding procedure can be computationally expensive.
We therefore also propose a delayed and intermittent update of the optimal arrival time of all the edges in the tree to decrease the computation complexity of the kinodynamic RRT* algorithm. 

The remainder of the paper is organized as follows. 
Some related works are given in Section~\ref{SecRelatedWorks}.
The statement of the problem studied in this paper is given in Section~\ref{SecProblemformulation}. 
In Section~\ref{SecPFF}, the PFF optimal controller is derived.
The PFF optimal controller is a key ingredient of the proposed Kino-RRT* algorithm, which is outlined in Section~\ref{SecKino-RRT*}. 
The implementation of Kino-RRT* on different robot systems is given in Section~\ref{SecExperiment}.
Finally, Section~\ref{secConclusion} concludes the paper.

\section{Related Works} \label{SecRelatedWorks}
Incremental sampling-based motion planning algorithms find an initial solution in high dimensional planning spaces quickly and then incrementally improving the solution.
For motion planning of robot systems, considering the differential constraints is necessary for generating feasible trajectories. 
The extension of RRT* to dynamic systems is studied in~\cite{Karaman2010Optimal}, where sufficient conditions ensuring asymptotic optimality of the RRT* for dynamic systems were established.
Every \textit{local steering} and \textit{distance function} in kinodynamic RRT* requires the solution of a TPBVP~\cite{Karaman2010Optimal}.
Assuming a solver of the TPBVP is available, references~\cite{karaman2013sampling,schmerling2015optimaldriftless,schmerling2015optimaldrift} study the radius of the neighbor nodes in kinodynamic RRT* to guarantee asymptotic optimality.

Solving the TPBVPs is the computationally expensive component of the kinodynamic RRT* algorithm.
Infinite-horizon and finite-horizon LQR controllers were used as the steering function in kinodynamic RRT* for linear or linearized systems in~\cite{Perez2012LQR} and~\cite{goretkin2013optimal}, respectively.
However, these methods cannot achieve the exact connection of two states, which is required in the kinodynamic RRT* algorithm.
A fixed-final-state free-final-time controller is used in~\cite{Dustin2013Kinodynamic} to achieve the exact connection of any pair of states for linear or linearized systems.
The optimal arrival time is computed by solving a root-finding problem.
To deal with nonlinear dynamics, \cite{Xie2015Towards} directly uses a numerical solver to solve the TPBVP online, and \cite{sakcak2019sampling} uses discrete motion primitives. 
Learning-based methods also have been studied to solve the TPBVP in kinodynamic RRT*.
References \cite{Wolfslag2018RRT-CoLearn} and \cite{zheng2021sampling} use offline generated optimal trajectories and supervised learning to train neural networks to solve the TPBVP.  
In~\cite{Chiang2019RLRRT}, the steering function is realized by a local policy trained using Deep Reinforcement Learning.

Other works solve the sampling-based kinodynamic motion planning problem without relying on TPBVP solvers ~\cite{hauser2016asymptotically, Li2016Asymptotically}.
These methods extended RRT-style shooting methods to kinodynamic planning by randomly sampling piece-wise constant control inputs of the system.
However, the convergence to high-quality trajectories in practice can be slow by the use of random controls~\cite{Xie2015Towards,Sivaramakrishnan2019,li2021mpc}.

\section{Problem formulation} \label{SecProblemformulation}

The optimal kinodynamic motion planning problem is defined as finding a dynamically feasible trajectory for the robot to reach the goal state starting from an initial state, while satisfying the state and control constraints and minimizing a cost function~\cite{Karaman2010Optimal,Dustin2013Kinodynamic}. 
Specifically, given the planning domain $X$, free space $X_\mathrm{free}$, goal region $X_\mathrm{goal}$, initial state $x_0$,  consider the dynamics of the robot 
\begin{equation}
\begin{split}
    \dot{x} =  A x + B u + c,
    \label{s3eq1}
\end{split}
\end{equation}
and the cost function,
\begin{equation}
\begin{split}
    J(u) = \int_{0}^{T} (1 + u^\top R u) \, \mathrm{d}\tau,
    \label{s3eq2}
\end{split}
\end{equation}
the goal of the motion planning problem is to find a control $u(t)$, $t \in [0,T]$, such that the solution $x(t)$ to (\ref{s3eq1}) is obstacle-free, i.e. $x(t) \in X_\mathrm{free}$, $t \in [0,T]$, reaches the goal region, i.e. $x(T) \in X_\mathrm{goal}$, and minimizes the cost functional (\ref{s3eq2}).
$A$, $B$, and $c$ are constant and given. (\ref{s3eq1}) represents the dynamics of a linear or linearized system.

RRT*-type algorithms try to solve this problem by growing a tree, which involves sampling intermediate states (nodes) and making optimal connections between states (edges).
This results in converging to the optimal solution asymptotically.
In kinodynamic RRT*, every edge between two states is the solution of a TPBVP given by
\begin{equation}
\begin{split}
u^*= & \mathop{\arg\min}_{u} \ J(u), \\
\mathrm{s.t.} \ \ & \dot{x}= A x + B u + c,\\
& x(0)=x_a, \ x(\tf)=x_b,
\label{s3eq3}
\end{split}
\end{equation}
where $J$ is the same as in (\ref{s3eq2}) but over the time interval $[0, \tf]$, and $x_a$ and $x_b$ are the sampled initial state and final state of this edge, respectively.
The solution of (\ref{s3eq3}) with free-final-time $\tf$ is given in~\cite{Dustin2013Kinodynamic}. 
Besides this fixed-final-state free-final-time controller, next, we will present a partial-final-state-free controller, which is the key ingredient of the proposed Kino-RRT* algorithm.

\section{Partial-Final-State-Free Optimal Controller} \label{SecPFF}

Rewrite the state $x \in \mathbb{R}^n$ as the concatenation of two vectors $x = [x_1^\top \ x_2^\top]^\top$, where $x_1 \in \mathbb{R}^{n_1}$ and $x_2 \in \mathbb{R}^{n_2}$ with $n_1 + n_2 = n$. 
The partial-final-state-free (PFF) optimal control problem is given by
\begin{equation}
\begin{split}
u^*= & \mathop{\arg\min}_{u} \ J(u), \\
\mathrm{s.t.} \ \ & \dot{x} = A x + B u + c,\\
& x(0)=x_a, \ x_1(\tf)=x_c.
\label{s4eq1}
\end{split}
\end{equation}
First, we consider the case where the arrival time $\tf$ is given.
Instead of fixing the states $x(0)$ and $x(\tf)$ as in (\ref{s3eq3}), only $x(0)$ and $x_1(\tf)$ are fixed, and $x_2(\tf)$ is free in (\ref{s4eq1}).

\subsection{The PFF Optimal Controller} \label{PFFController}
We solve this PFF optimal control problem using Pontryagin's Maximum Principle~\cite{lewis2012optimal}. 
The Hamiltonian of the system is given by
\begin{equation}
    H(x,u,t,\lambda) = 1 + u^\top R u + \lambda^\top (Ax + Bu + c).
    \label{eqHamilt}
\end{equation}
The necessary conditions for optimality are
\begin{align}
    \dot{x} &= Ax +Bu + c, \label{eq4} \\ 
    \dot{\lambda} &= - \frac{\partial H}{\partial x} = - A^\top \lambda, \label{eq5} \\
    0 &= \frac{\partial H}{\partial u} = 2 R u + B^\top \lambda, \label{eq6} \\
    \mathrm{0} &= \lambda_2(\tf), \label{eq7}
\end{align}
where $\lambda = [\lambda_1^\top \ \lambda_2^\top]^\top$, $\lambda_1 \in \mathbb{R}^{n_1} $ is the costate of $x_1$, and $\lambda_2 \in \mathbb{R}^{n_2}$ is the costate of $x_2$.
Solving for $u$ using (\ref{eq6}), we get
\begin{equation}
    u = - \frac{1}{2}R^{-1} B^\top \lambda.
    \label{equ}
\end{equation}
Substituting (\ref{equ}) into (\ref{eq4}), yields
\begin{equation}
    \dot{x} = Ax - \frac{1}{2} B R^{-1} B^\top \lambda + c.
    \label{eq8}
\end{equation}
The analytical solutions for the differential equations (\ref{eq5}) and (\ref{eq8}) are available and are given by
\begin{align}
    \lambda(t) &= e^{A^\top (\tf-t)} \lambda(\tf), \label{eq9} \\ 
    x(t) &= e^{At} x(0) - \frac{1}{2} G(t) \lambda(\tf) + \int_0^t e^{A(t-\tau)}c \, \mathrm{d}\tau, \label{eq10}
\end{align}
where $G(t) = \int_0^t e^{A(t- \tau)} B R^{-1} B^\top e^{A^\top (\tf - \tau)} \, \mathrm{d}\tau$.

Note that if $\lambda(\tf)$ is known, then the problem can be solved with the control given by (\ref{equ}) and (\ref{eq9}), and the state trajectory given by (\ref{eq10}).
Thus, the problem remains to determine $\lambda_1(\tf)$.
To this end, evaluate (\ref{eq10}) at $\tf$ to obtain 
\begin{equation}
     x(\tf) = \bar{x}(\tf) - \frac{1}{2} G(\tf) \lambda(\tf), \label{eq11}
\end{equation}
where
\begin{equation}
    \bar{x}(\tf) \triangleq e^{A \tf} x(0) + \int_0^\tf e^{A(\tf-\tau)}c \, \mathrm{d}\tau.
\end{equation}
We may obtain $x_2(\tf)$ and $\lambda_1(\tf)$ by solving the linear equations (\ref{eq11}).
Using (\ref{eq7}), rewrite (\ref{eq11}) as
\begin{equation}
     \begin{bmatrix} \bar{x}_1(\tf) - x_1(\tf) \\ \bar{x}_2(\tf) - x_2(\tf) \end{bmatrix} = \frac{1}{2}\begin{bmatrix} G_{11}(\tf) & G_{12}(\tf)  \\ G_{21}(\tf) & G_{22}(\tf) \end{bmatrix} \begin{bmatrix} \lambda_1(\tf) \\ \mathrm{0} \end{bmatrix},
     \label{eq12}
\end{equation}
where $\bar{x}(\tf) = [\bar{x}_1^\top(\tf) \ x_2^\top(\tf)]^\top$.
Note that $\bar{x}_1(\tf) - x_1(\tf)$ is known and $\bar{x}_2(\tf) - x_2(\tf)$ is unknown.
Then, (\ref{eq12}) becomes
\begin{align}
    2 (\bar{x}_1(\tf) - x_1(\tf)) &= G_{11}(\tf) \lambda_1(\tf), \label{eq13}\\ 
    2 (\bar{x}_2(\tf) - x_2(\tf)) &= G_{21}(\tf) \lambda_1(\tf). \label{eq14}
\end{align}
Assuming $(A, B)$ is controllable, it follows that $G(\tf)$ is invertible and hence $G_{11}(\tf)$ is invertible.
From (\ref{eq13}), we can solve for  $\lambda_1(\tf)$ as follows
\begin{equation}
    \lambda_1(\tf) = 2 G_{11}^{-1}(\tf) (\bar{x}_1(\tf) - x_1(\tf)).
\end{equation}
$x_2(\tf)$ can be computed from (\ref{eq14}). 
Finally, from (\ref{equ}) and (\ref{eq9}), the open-loop optimal control is given by
\begin{equation}
     u(t) = - \frac{1}{2} R^{-1} B^\top e^{A^\top (\tf-t)} \lambda(\tf). \label{eq15}
\end{equation}
Substituting (\ref{eq15}) into (\ref{s3eq2}), the optimal cost is 
\begin{equation}
    J(u^*) = \tf + \frac{1}{4} \lambda(\tf)^\top G(\tf) \lambda(\tf).
\end{equation}

\subsection{The Optimal Arrival Time}
Next, consider the case when $\tf$ is free.
In this case, we have the transversality condition~\cite{lewis2012optimal}
\begin{equation}
     H(\tf) = 0. \label{eq16}
\end{equation}
Substituting (\ref{equ}) into (\ref{eqHamilt}) and evaluating (\ref{eqHamilt}) at $\tf$, then (\ref{eq16}) becomes
\begin{equation}
\begin{split}
     H(\tf) = 1 + \lambda(\tf)^\top (A x(\tf) + c) - \frac{1}{4} \lambda(\tf)^\top B R^{-1} B^\top \lambda(\tf) = 0.
     \label{eqHamiTf}
\end{split}
\end{equation}
We find the optimal arrival time $\tf$ by solving (\ref{eqHamiTf}), which requires finding the roots of a polynomial~\cite{Dustin2013Kinodynamic}. 

\subsection{PFF with Quadratic Terminal Penalty} \label{PFFPenalty}
In some cases, it may be desired to add implicit constraints on the free-final-state.
Here, we extend the PFF optimal controller by adding a quadratic penalty on the free-final-state to the cost function.
Consider the PFF optimal control problem with the cost function,
\begin{equation}
\begin{split}
    J(u) = \frac{1}{2} x_2(\tf)^\top S x_2(\tf) + \int_{0}^{\tf} (1 + u^\top R u) \, \mathrm{d}\tau.
\end{split}
\end{equation}
The necessary conditions for optimality for the PFF control problem (\ref{s4eq1}) with this new cost function are still given by (\ref{eq4})-(\ref{eq6}), except that (\ref{eq7}) is now replaced by
\begin{align}
    \lambda_2(\tf) &= \phi_x^\top(x(\tf)) = S x_2(\tf), \label{s4eq24}
\end{align}
where $ \phi(x(\tf)) = \frac{1}{2} x_2(\tf)^\top S x_2(\tf)$.

Following the same derivation as before, we get the same expression given by (\ref{eq11}). 
The problem remains to solve for $\lambda(\tf)$. 
Substituting (\ref{s4eq24}) into (\ref{eq11}), we get
\begin{equation}
     \bar{x}(\tf) - x(\tf) = \frac{1}{2}G(\tf) \begin{bmatrix} \lambda_1(\tf) \\ S x_2(\tf) \end{bmatrix},
\end{equation}
which is equvalent to
\begin{equation}
    \begin{bmatrix} \bar{x}_1(\tf) - x_1(\tf) \\ \bar{x}_2(\tf) \end{bmatrix} = M \begin{bmatrix} \lambda_1(\tf) \\ x_2(\tf) \end{bmatrix}, \label{s4eq26}
\end{equation}
where 
\begin{equation}
    M = \left( \frac{1}{2}G(\tf) \begin{bmatrix} I & \mathrm{0} \\ \mathrm{0} & S \end{bmatrix} + \begin{bmatrix} \mathrm{0} & \mathrm{0} \\ \mathrm{0} & I \end{bmatrix} \right). 
\end{equation}
Note that $M$ is invertible.
Thus, we can calculate $\lambda_1(\tf)$ and $x_2(\tf)$ from (\ref{s4eq26}).
Along with (\ref{s4eq24}), we obtain $\lambda(\tf)$.

\section{The Kino-RRT* Algorithm} \label{SecKino-RRT*}

In this section, we present the details of the Kino-RRT* algorithm, which is built on both the PFF controller and the fixed-final-state free-final-time controller.
First, we summarize some primitive procedures used in the Kino-RRT* algorithm.
Some of these primitive procedures follow the work in~\cite{Karaman2011Sampling}. \\
\textbf{Sampling:} The sampling procedure \texttt{SamplePFF} returns a partial state that is randomly sampled in a reduced state space and is collision-free in the corresponding reduced state space. 
For example, for a robot whose state space includes the position space and the velocity space, \texttt{SamplePFF} may sample a position of the robot that is collision-free.  \\
\textbf{Parent:} \texttt{parent}$(x)$ returns the parent node of $x$.  \\
\textbf{Nearest Neighbor:} Given a tree $G = (V,E)$, where $V$ is the node set and $E$ is the edge set, the procedure \texttt{Nearest}$(V,x)$ returns the node in $V$ that is closest to the state $x$. \\ 
\textbf{Near Nodes:} The function \texttt{Near}$(V,x,r)$ returns all the nodes in $V$ that are contained in a ball of radius $r$ centered at $x$. \\ 
\textbf{Collision Checking:} The function \texttt{CollisionFree}$(\tau)$ takes a trajectory $\tau$ (an edge segment) as an input and returns true if and only if $\tau$ lies entirely in the collision-free space.
The function \texttt{CollisionPoint}$(x)$ returns true if and only if the point $x$ is collision-free. \\
\textbf{Cost:} The procedure \texttt{Cost}$(x)$ returns the cost-to-come from the root node to $x$. \\
\textbf{Segment Cost:} The procedure \texttt{SegCost}$(x_i,x_j)$ returns the cost to go from $x_i$ to $x_j$.
Depending on $x_j$, this cost is obtained by either solving the PFF control problem or the fixed-final-state free-final-time control problem. \\
\textbf{Shrink:} The procedure \texttt{Shrink}$(x_i, x_j)$ returns $x_j$ if the distance between $x_i$ and $x_j$ is less than or equal to $\ell$.
Otherwise, it returns a new state $x_\mathrm{new}$ that lies on the line formed by $x_i$ and $x_j$ and is at a distance $\ell$ away from $x_i$ towards $x_j$.
The \texttt{Shrink} procedure is consistent with the RRT* algorithm dictates that segments should have a maximum length $\ell$.
If one tries to connect two points that are far away, this connecting segment will collide with obstacles with a high probability. \\
\textbf{Steering:} The procedure \texttt{SteerPFF}$(x_i,x_j)$ solves the TPBVP using the PFF optimal controller, and it returns a trajectory $\tau$ that starts from $x_i$ and ends at $x_j$.
The procedure \texttt{Steer}$(x_i,x_j)$ solves the TPBVP using the fixed-final-state free-final-time controller, and it returns a trajectory $\tau$ that starts from $x_i$ and ends at $x_j$. 
Note that $x_j$ in \texttt{Steer}$(x_i,x_j)$ is a point in the full state space, while $x_j$ in \texttt{SteerPFF}$(x_i,x_j)$ is a point in the reduced sampling space. \\
\textbf{FreeState:} The function \texttt{FreeSate} takes the trajectory $\tau$ returned by \texttt{SteerPFF}$(x_i,x_j)$ as input and returns the rest of the state $x_\mathrm{free}$ at the endpoint of the trajectory that is not specified by $x_j$. 

\IncMargin{.5em}
\begin{algorithm}
\caption{Kino-RRT*}
\label{alg:Kino-RRT*}

$V \leftarrow \{ x_\mathrm{init} \}$; $E \leftarrow \emptyset$; $G \leftarrow (V,E)$\;
\For{$i=1:N$} 
{
    $z_\mathrm{rand} \leftarrow \texttt{SamplePFF}$\;
    $x_\mathrm{nearest} \leftarrow \texttt{Nearest}(V,z_\mathrm{rand})$\;
    $z_\mathrm{new} \leftarrow \texttt{Shrink} (x_\mathrm{nearest}, z_\mathrm{rand})$\;
    \If {$\texttt{CollisionPoint}(z_\mathrm{new})$}
    {
        $\tau \leftarrow \texttt{SteerPFF}(x_\mathrm{nearest},z_\mathrm{new})$\;
        \If{$\texttt{CollisionFree}(\tau)$}
        {   
            $x_\mathrm{free} \leftarrow \texttt{FreeState}(\tau)$\;
            $X_\mathrm{near} \leftarrow \texttt{Near}(V,z_\mathrm{new}, r)$\;
            $(x_\mathrm{min},x_\mathrm{free}) \leftarrow \texttt{ChooseParent}(X_\mathrm{near},x_\mathrm{nearest},z_\mathrm{new})$\;
            $x_\mathrm{new} \leftarrow (z_\mathrm{new}, x_\mathrm{free})$\;
            $V \leftarrow V \cup \{ x_\mathrm{new} \}$\;
            $E \leftarrow E \cup \{ (x_\mathrm{min},x_\mathrm{new}) \}$\;
            $E \leftarrow \texttt{Rewire}(X_\mathrm{near},E,x_\mathrm{new},x_\mathrm{min})$\;
            $G \leftarrow (V,E)$;
         }
    }
}
\KwRet $G$;
\end{algorithm}
\DecMargin{.5em}

\IncMargin{.3em}
\begin{algorithm}
\caption{ChooseParent}
\label{alg:ChooseParent}
\SetKwFunction{ChooseParent}{}
\SetKwProg{Fn}{ChooseParent}{:}{}
\Fn{\ChooseParent{$X_\mathrm{near},x_\mathrm{nearest},z_\mathrm{new}$}}
{
    $x_\mathrm{min}\leftarrow x_\mathrm{nearest}$\;
    $c_\mathrm{min} \leftarrow \texttt{Cost}(x_\mathrm{nearest})+\texttt{SegCost}(x_\mathrm{nearest},z_\mathrm{new})$\;
    \ForEach{$x_\mathrm{near} \in X_\mathrm{near} \setminus {x_\mathrm{nearest}}$} 
    {
        \If {$\texttt{Cost}(x_\mathrm{near})+\texttt{SegCost}(x_\mathrm{near},z_\mathrm{new}) < c_\mathrm{min}$}
        {
            $\tau \leftarrow \texttt{SteerPFF}(x_\mathrm{near},z_\mathrm{new})$\;
            \If{$\texttt{CollisionFree}(\tau)$}
            {
                $x_\mathrm{free} \leftarrow \texttt{FreeState}(\tau)$\;
                $x_\mathrm{min}\leftarrow x_\mathrm{near}$\;
                $c_\mathrm{min} \leftarrow \texttt{Cost}(x_\mathrm{near})+\texttt{SegCost}(x_\mathrm{near},z_\mathrm{new})$\;
            }
        }
    }
    \KwRet $(x_\mathrm{min},x_\mathrm{free})$\;
}
\end{algorithm}
\DecMargin{.3em}

\IncMargin{.5em}
\begin{algorithm}
\caption{Rewire}
\label{alg:Rewire}
\SetKwFunction{Rewire}{}
\SetKwProg{Fn}{Rewire}{:}{}
\Fn{\Rewire{$X_\mathrm{near},E,x_\mathrm{new},x_\mathrm{min}$}}
{
    \ForEach{$x_\mathrm{near} \in X_\mathrm{near} \setminus {x_\mathrm{min}}$} 
    {
        \If {$\texttt{Cost}(x_\mathrm{new})+\texttt{SegCost}(x_\mathrm{new},x_\mathrm{near}) < \texttt{Cost}(x_\mathrm{near})$}
        {
            $\tau \leftarrow \texttt{Steer}(x_\mathrm{new},x_\mathrm{near})$\;
            \If{$\texttt{CollisionFree}(\tau)$}
            {
                $x_\mathrm{parent}\leftarrow \texttt{Parent}(x_\mathrm{near})$\;
                $E \leftarrow E \setminus \{ (x_\mathrm{parent},x_\mathrm{near}) \}$\;
                $E \leftarrow E \cup \{ (x_\mathrm{new},x_\mathrm{near}) \}$\;
            }
        }
    }
    \KwRet $E$\;
}
\end{algorithm}
\DecMargin{.5em}

The complete algorithm is given by Algorithm \ref{alg:Kino-RRT*}, Algorithm \ref{alg:ChooseParent}, and Algorithm \ref{alg:Rewire}.
We use $z$ to denote a point in the reduced sampling space.
The rest of the state (free-state) $x_\mathrm{free}$, which comes from the endpoint of the edge segment (state trajectory), is decided by the PFF optimal controller.
After the \texttt{ChooseParent} step (line 11, Algorithm \ref{alg:Kino-RRT*}), the free-state is found and is combined with the sampled state to form a point in the full state space (line 12, Algorithm \ref{alg:Kino-RRT*}).
Then, this point is added to the tree as a node (line 13, Algorithm \ref{alg:Kino-RRT*}).

\subsection{Delayed and Intermittent Update of the Arrival Time}
\begin{figure}[htb]
    \centering
    \begin{subfigure}[b]{0.49\columnwidth}
         \centering
         \includegraphics[width=1\columnwidth]{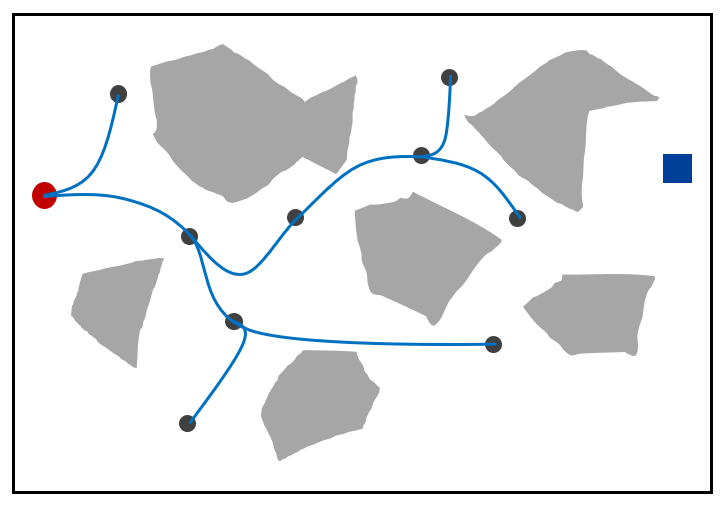}
         \caption{}
     \end{subfigure}
     \begin{subfigure}[b]{0.49\columnwidth}
         \centering
         \includegraphics[width=1\columnwidth]{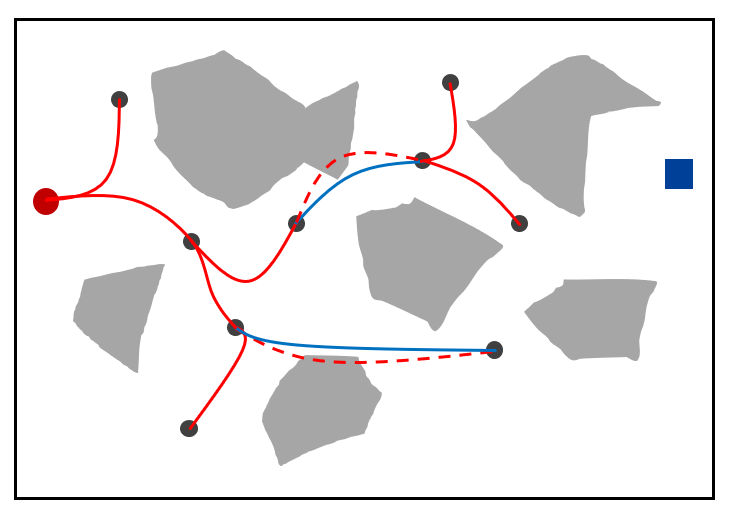}
         \caption{}
     \end{subfigure}
        \caption{Delayed and intermittent update of the arrival time. (a) Grow a tree using a heuristic of the arrival time (blue lines). (b) Delayed update of the optimal arrival time (red lines). If the updated edge is in-collision (red dash lines), the original edge is used (blue lines). }
        \label{DelayedUpdate}
\end{figure}

For both the PFF controller and the fixed-final-state controller, finding the optimal arrival time of the TPBVP requires solving a root-finding problem of a high-order polynomial (see (\ref{eqHamiTf})).
This root-finding procedure will slow down the kinodynamic RRT* algorithm, as the TPBVP is required to be solved repeatedly.
Here we propose a delayed and intermittent update of the optimal arrival time, which is shown in Figure~\ref{DelayedUpdate}.
The planning algorithm first grows a tree using a heuristic of the arrival time (for example, by setting a desired average speed) without solving the root-finding problem (Figure~\ref{DelayedUpdate}(a)). 
Then, we intermittently update all the edges in the tree using the optimal arrival time (Figure~\ref{DelayedUpdate}(b)). 
If the updated edge is in-collision, we will use the original edge.
We call this method KinoD-RRT*. 

\section{Experimental Results} \label{SecExperiment}
We tested the Kino-RRT* algorithm on two kinodynamic systems: a 2D double integrator robot operating in a plane environment and a linearized quadrotor robot with a 10-dimensional state-space.
We compared the Kino-RRT* algorithm with a variant of the kinodynamic RRT* algorithm.
The only difference between the Kino-RRT* and the compared algorithm (a variant of kinodynamic RRT*) is the utilization of the PFF controller in Kino-RRT*. 
The compared kinodynamic RRT* algorithm samples the full state space and uses the fixed-final-state free-final-time controller to solve the TPBVPs.
The gain of performance is solely due to the PFF controller.
Thus, this comparison is informative.

\subsection{Implementation Details}

In kinodynamic RRT*, the near nodes are found by using the forward-reachable set or the backward-reachable set~\cite{Dustin2013Kinodynamic,schmerling2015optimaldrift}.
Specifically, in line 12, Algorithm \ref{alg:Kino-RRT*}, $\texttt{Near}(V,x, r)$ returns all nodes in $V$ such that the cost $J$ to go from these nodes to $x$ is less than $r$ (backward-reachable set).
Check membership in the forward/backward reachable set for a set of nodes can be computationally expensive. 

We use Euclidean distance to find the near nodes and the nearest node.
This essentially means that we do not use the true distance. 
In this case, the forward-reachable set and the backward-reachable set are the same.
For kinodynamic motion planning, the true distance between two states is the minimum cost $J$ from the solution of the TPBVP~\cite{Karaman2010Optimal}. 
Using the true distance, the forward (or backward) reachable set defines an $\epsilon$-radius sub-Riemannian ball centered at $x$.
It is showed in~\cite{Li2016Asymptotically} that there always exists a certain size Euclidean hyper-ball inside such sub-Riemannian ball under mild conditions, which justifies the use of Euclidean norms. 
Euclidean distance is also used in~\cite{Li2016Asymptotically}.
After the nearest node and the near nodes are selected, the true distance is used in the \texttt{ChooseParent} and \texttt{Rewire} algorithms.
The Euclidean distance is used only to pre-select relevant nodes and to help with the computations.

We also used a constant radius for the Euclidean hyper-ball for the near nodes, which implies a constant radius of the sub-Riemannian ball with respect to the true distance.
Note that the kinodynamic RRT* is asymptotically optimal with a constant neighbor radius.
The implementation is the same for the Kino-RRT* and the compared algorithm for an informative comparison.
All experiments are done on a laptop computer with an Intel Core i5-8250U 1.6 GHz CPU and 8 GB of RAM.

\subsection{2D Double Integrator} \label{SubSec2DDI}

The state of the 2D double integrator is given by $x = [p^\top \ v^\top]^\top$, where $p = [x_1 \ x_2]^\top$ is the position and $v = [x_3 \ x_4]^\top$ is the velocity.
The control input is the acceleration.
The system matrices are given by 
\begin{equation}
    A = \begin{bmatrix} 0 & I_2 \\ 0 & 0 \end{bmatrix}, \quad
    B = \begin{bmatrix} 0 \\ I_2 \end{bmatrix}, \quad
    c = 0.
\end{equation}
The weighting matrix in the cost function is set to $R = I_2$. 

For both Kino-RRT* and kinodynamic RRT* the position is uniformly sampled within the boundary of the environment, that is, $p \in [0, 20]^2 \ \mathrm{m}$.
The free-final-state of the PFF controller is the velocity.
Thus, the Kino-RRT* algorithm does not sample the velocity space.
For the kinodynamic RRT* algorithm, the velocity is uniformly sampled in $v \in [-2, 2]^2 \ \mathrm{m/s^2}$.  
Note that a larger interval for the velocity essentially requires searching in a larger state space, which will result in slower convergence.
However, if the sampling velocity interval is too small, the search is confined to a small state space that may not contain the optimal solution.
Here, the velocity interval is chosen to be small while containing the optimal solution.  

\begin{figure}
    \centering
    \begin{subfigure}[b]{0.49\columnwidth}
         \centering
         \includegraphics[width=1\columnwidth]{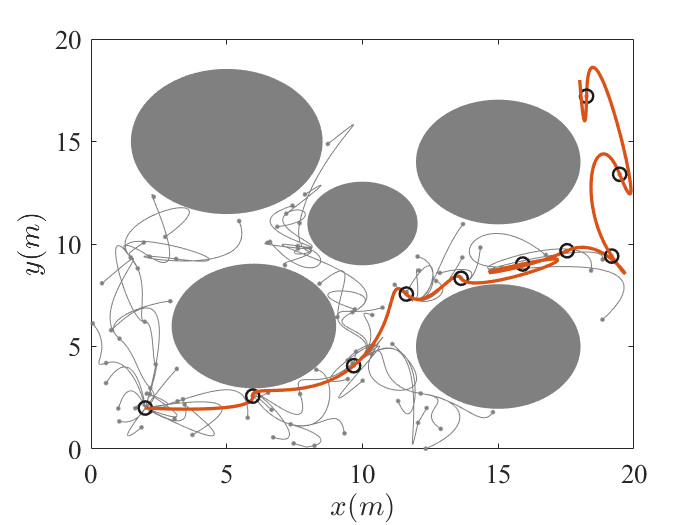}
     \end{subfigure}
     \begin{subfigure}[b]{0.49\columnwidth}
         \centering
         \includegraphics[width=1\columnwidth]{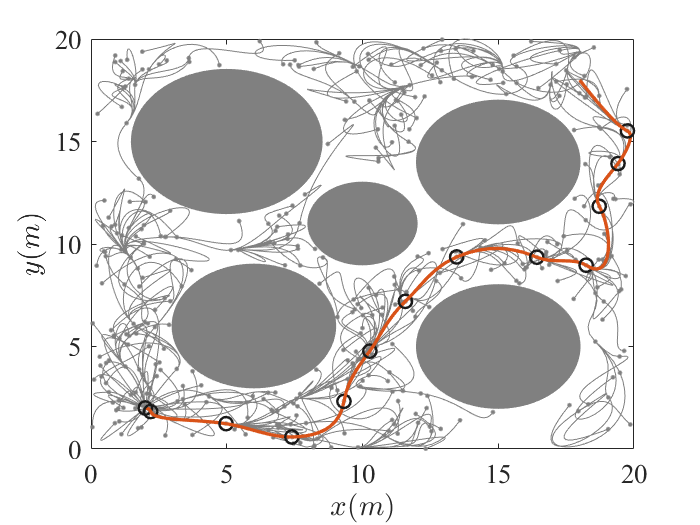}
     \end{subfigure}
     \begin{subfigure}[b]{0.49\columnwidth}
         \centering
         \includegraphics[width=1\columnwidth]{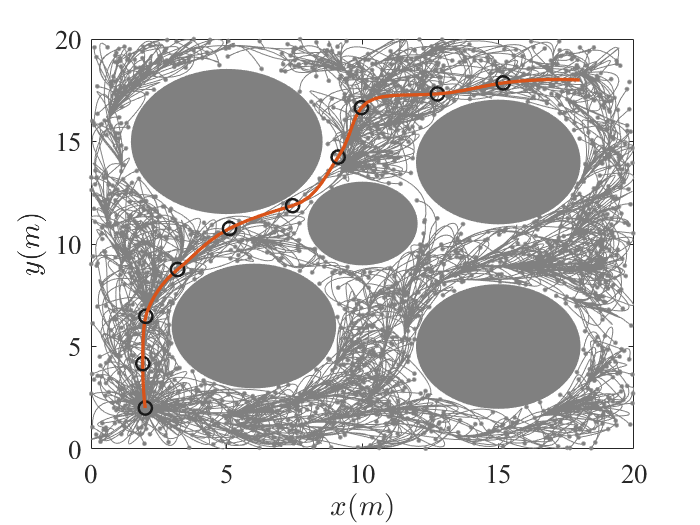}
     \end{subfigure}
     \begin{subfigure}[b]{0.49\columnwidth}
         \centering
         \includegraphics[width=1\columnwidth]{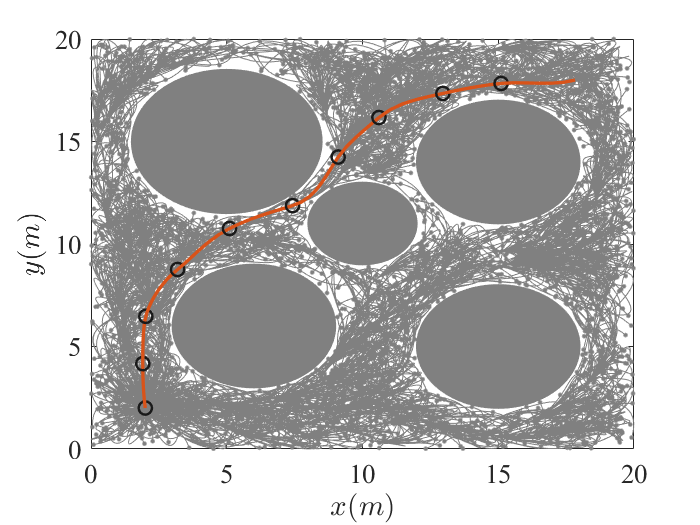}
     \end{subfigure}
        \caption{Kinodynamic RRT* results of the 2D double integrator. The first figure corresponds to the first solution found. From the upper left to bottom right, the nodes expanded are $94$, $400$, $2000$, $4000$. The corresponding time to generate these trees are $0.053$, $0.22$, $2.38$, $8.17$ sec. The cost of best trajectory in the trees are $86.76$, $47.56$, $27.99$, $25.72$.}
        \label{TreeDIcomp1}
\end{figure}

\begin{figure}
    \centering
    \begin{subfigure}[b]{0.49\columnwidth}
         \centering
         \includegraphics[width=1\columnwidth]{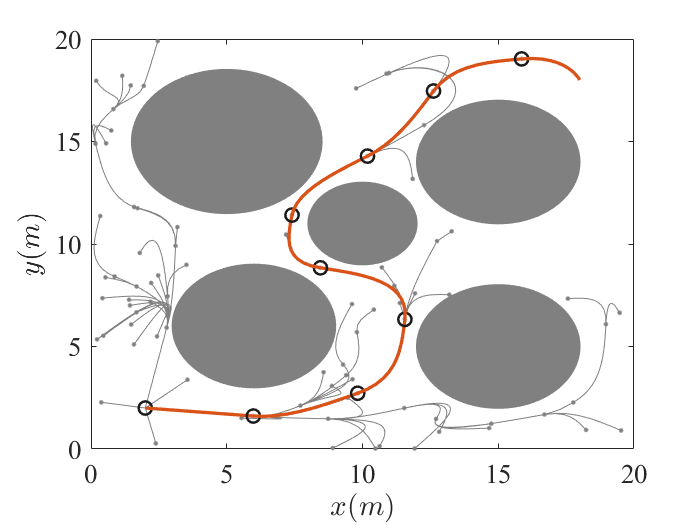}
     \end{subfigure}
     \begin{subfigure}[b]{0.49\columnwidth}
         \centering
         \includegraphics[width=1\columnwidth]{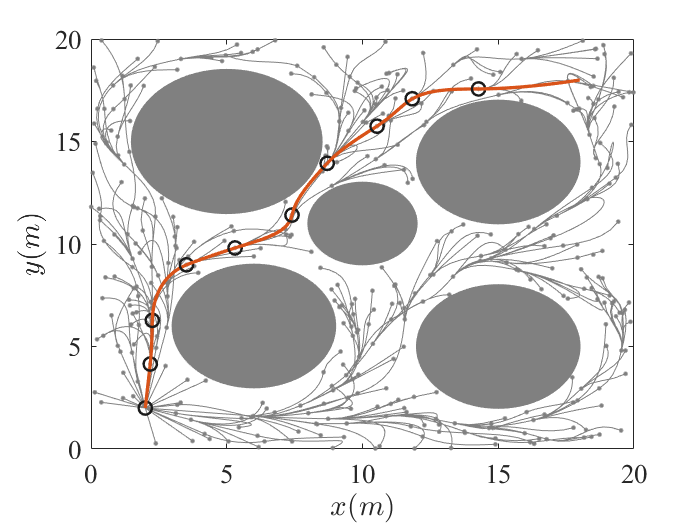}
     \end{subfigure}
     \begin{subfigure}[b]{0.49\columnwidth}
         \centering
         \includegraphics[width=1\columnwidth]{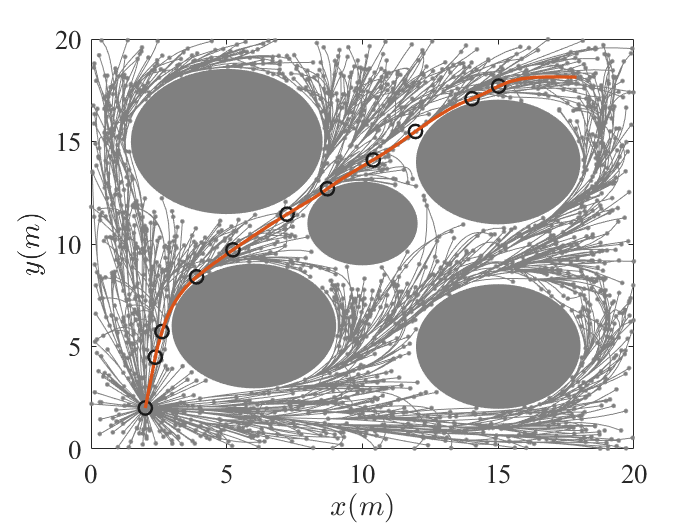}
     \end{subfigure}
     \begin{subfigure}[b]{0.49\columnwidth}
         \centering
         \includegraphics[width=1\columnwidth]{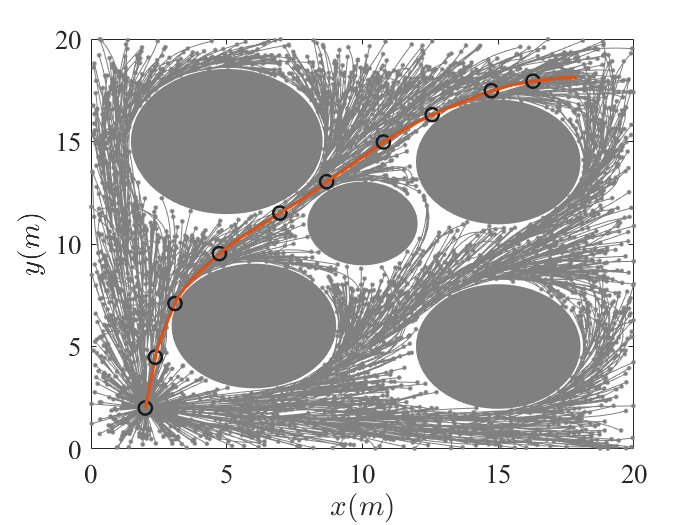}
     \end{subfigure}
        \caption{Kino-RRT* results of the 2D double integrator. The first figure corresponds to the first solution found. From the upper left to bottom right, the nodes in the tree are $85$, $400$, $2000$, $4000$. The corresponding time to generate these trees are $0.015$, $0.14$, $2.12$, $7.64$ sec. The cost of best trajectory in the trees are $37.46$, $25.97$, $18.72$, $16.42$.}
        \label{TreeDIcomp2}
\end{figure}

\begin{figure}
    \centering
    \includegraphics[width=0.55\columnwidth]{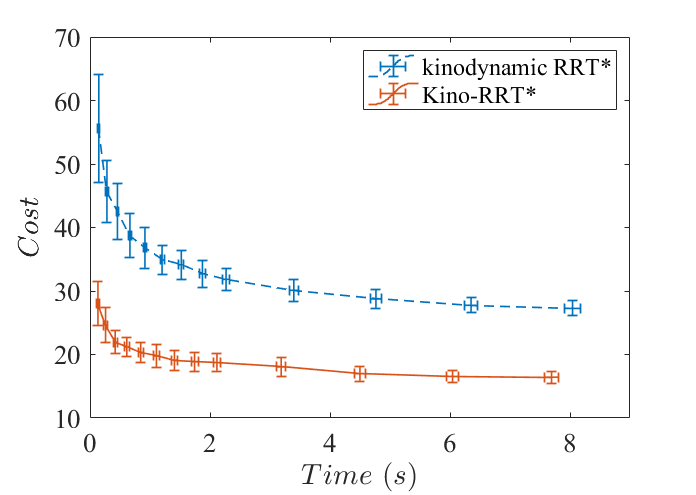}
        \caption{Comparison of Kino-RRT* and kinodynamic RRT* for the 2D double integrator case.}
        \label{CostDIcomp}
\end{figure}

The results of the kinodynamic RRT* algorithm and the Kino-RRT* algorithm are given in Figure~\ref{TreeDIcomp1} and Figure~\ref{TreeDIcomp2}, respectively.
The comparison of the Kino-RRT* and the kinodynamic RRT* is shown in Figure~\ref{CostDIcomp}. 
In Figure~\ref{CostDIcomp}, we can see that our algorithm finds a better trajectory from the beginning (the first solution).
In fact, the solution found by Kino-RRT* within 0.14 sec is comparable to the solution found by kinodynamic RRT* that took 8 sec after expanding 4000 nodes.
After the Kino-RRT* finds the first solution, the cost enters a sharp decrease phase.
For the kinodynamic RRT* algorithm, the cost curve is close to flat after 8 sec, and the probability of sampling good states to decrease the cost is low.
Kino-RRT* is more than 50 times faster than the kinodynamic RRT* to find a trajectory with the same cost.
By sampling in a reduced state-space, the solution returned by Kino-RRT* is close to the optimal solution after a few seconds of computation.
However, for the kinodynamic RRT* algorithm, it is difficult to sample good velocities that are comparable to the ones chosen by the PFF optimal controller, which leads to slow convergence. 

\begin{figure}
    \centering
    \begin{subfigure}[b]{0.49\columnwidth}
         \centering
         \includegraphics[width=1\columnwidth]{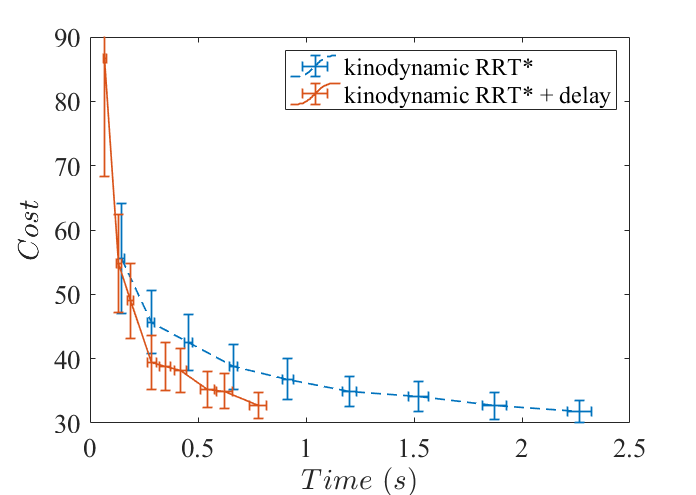}
         \caption{}
     \end{subfigure}
     \begin{subfigure}[b]{0.49\columnwidth}
         \centering
         \includegraphics[width=1\columnwidth]{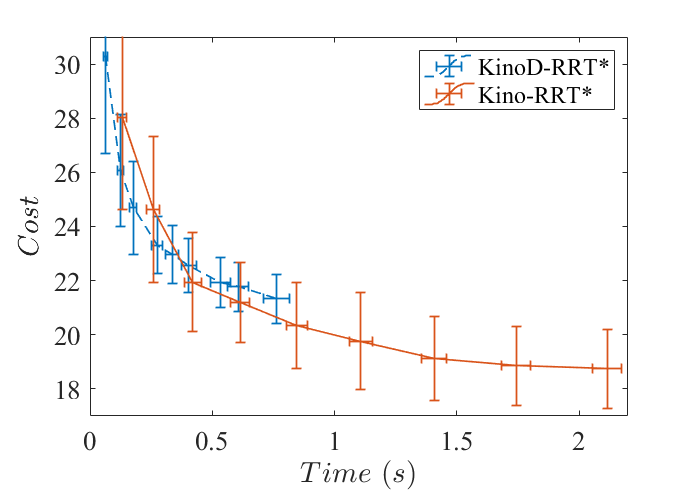}
         \caption{}
     \end{subfigure}
        \caption{Delayed and intermittent update of the optimal arrival time of the 2D double integrator. The arrival time is updated whenever another 500 nodes are added to the tree.}
        \label{DelayedDIcomp}
\end{figure}

Figure~\ref{DelayedDIcomp} shows the results of the delayed and intermittent update of the optimal arrival time. 
The Kinodynamic RRT* combined with the delayed and intermittent update of the optimal arrival time is called Kinodynamic RRT* with delay.
Four methods, Kinodynamic RRT*, Kinodynamic RRT* with delay, Kino-RRT*, and KinoD-RRT*, are compared.
Kinodynamic RRT* with delay is 3 times faster than Kinodynamic RRT* when expanding the same number of nodes.
The planned trajectories have a similar cost for expanding the same number of nodes.

Kino-RRT* with delay is also 3 times faster than Kino-RRT* when expanding the same number of nodes. 
We can see that in Figure~\ref{DelayedDIcomp}(b), KinoD-RRT* (blue dash line) finds a better trajectory in the beginning because it can expand more nodes in a given time.
However, Kino-RRT* outperforms KinoD-RRT* after some point.
This is because the velocities (free-final-state) chosen by KinoD-RRT* are not as optimized as the velocities chosen by Kino-RRT*.
The velocity chosen by the PFF controller is affected by the arrival time.
Non-optimal arrival times (which is the case with KinoD-RRT*) will result in a sub-optimal final velocity.
Thus, the performance of delayed update depends on the heuristic for the arrival time.

\subsection{Linearized Quadrotor}

A linearized quadrotor model adopted from~\cite{Dustin2013Kinodynamic} is used.
The 10-dimensional state is given by $x = [p^\top \ v^\top \ r^\top \ w^\top]^\top$, which consists of the three-dimensional position $p$ and velocity $v$, and the two-dimensional orientation $r$ and angular velocity $w$.
The yaw rotation, which is a redundant degree of freedom, is not considered in the model.
The system matrices are given by 
\begin{align*}
    A &= \begin{bmatrix} 0 & I_3 & 0 & 0 \\ 0 & 0 & \begin{bmatrix} 0 & g \\ -g & 0 \\ 0 & 0 \end{bmatrix} & 0 \\ 0 & 0 & 0 & I_2 \\ 0 & 0 & 0 & 0\end{bmatrix}, \quad
    B = \begin{bmatrix} 0 & 0 \\ \begin{bmatrix} 0 \\ 0 \\  \frac{1}{m} \end{bmatrix} & 0 \\ 0 & 0 \\ 0 & \frac{\ell I_2}{J} \end{bmatrix}, \\
    c &= 0, 
\end{align*}
where $g$ is the gravitational acceleration, $m$ is the mass of the quadrotor, $\ell$ is the distance between the center of the vehicle and the rotors, and $J$ is the moment of inertia about the axes coplanar with the rotors. 
The control input of the system is $u = [u_f \ u_x \ u_y]^\top$, where $u_f$ is the total thrust of the rotors relative to the thrust needed for hovering, and $u_x$ and $u_y$ are the relative torques of roll and pitch, respectively.

The free-final-state of the PFF controller is $v$, $r$, and $w$.
Thus the Kino-RRT* algorithm only samples the position space.
Since the quadrotor is linearized at the hovering state and the dynamics is sensitive to the roll and pitch angles, we will use the PPF controller with quadratic terminal penalty introduced in Section~\ref{PFFPenalty}.
The terminal penalty matrix is $S = \mathrm{diag}(0, 0, 0, 20, 20, 0, 0)$.
The weighting matrix of the control is $R = \mathrm{diag}(15, 30, 30)$. 

For both Kino-RRT* and kinodynamic RRT* the position is uniformly sampled within the boundary of the 3D environment.
The sampling intervals of $v$, $r$, and $w$ for the kinodynamic RRT* are $v \in [-2, 2]^3 \ \mathrm{m/s}$, $r \in [-1, 1]^2 \ \mathrm{rad}$, and $w \in [-4, 4]^2 \ \mathrm{rad/s}$, respectively.  

\begin{figure}
    \centering
    \begin{subfigure}[b]{0.49\columnwidth}
         \centering
         \includegraphics[width=1\columnwidth]{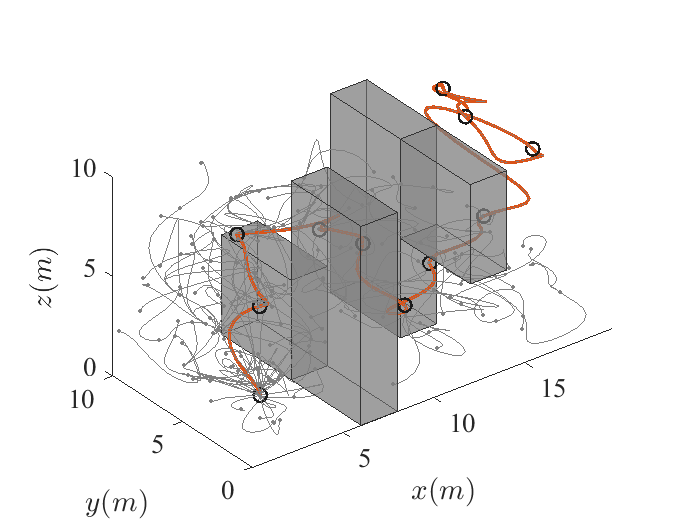}
     \end{subfigure}
     \begin{subfigure}[b]{0.49\columnwidth}
         \centering
         \includegraphics[width=1\columnwidth]{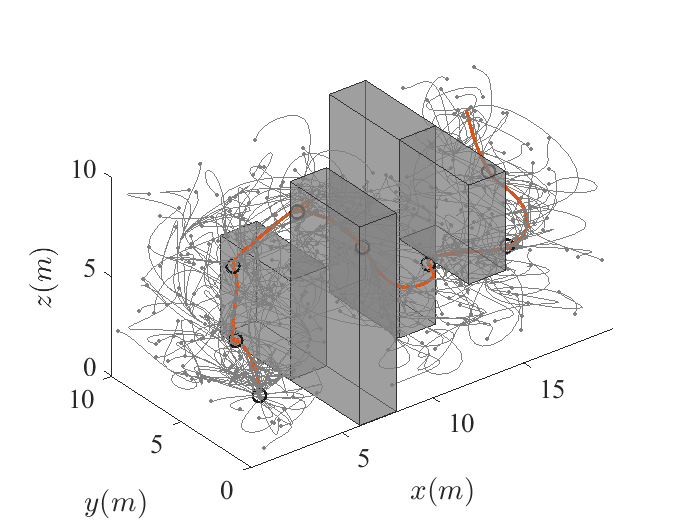}
     \end{subfigure}
     \begin{subfigure}[b]{0.49\columnwidth}
         \centering
         \includegraphics[width=1\columnwidth]{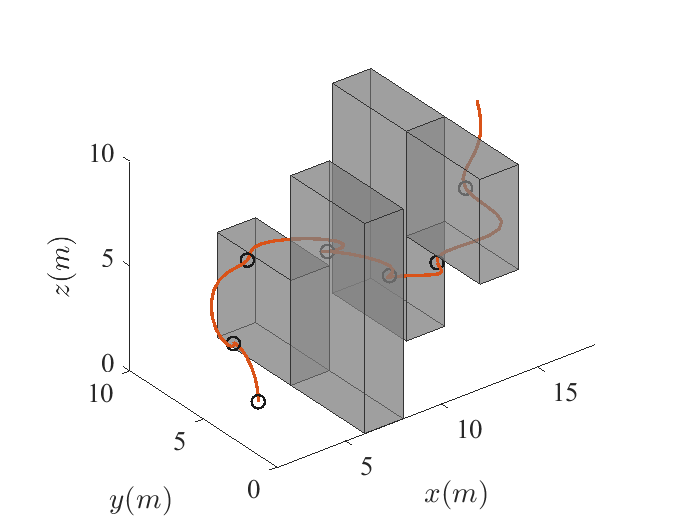}
     \end{subfigure}
     \begin{subfigure}[b]{0.49\columnwidth}
         \centering
         \includegraphics[width=1\columnwidth]{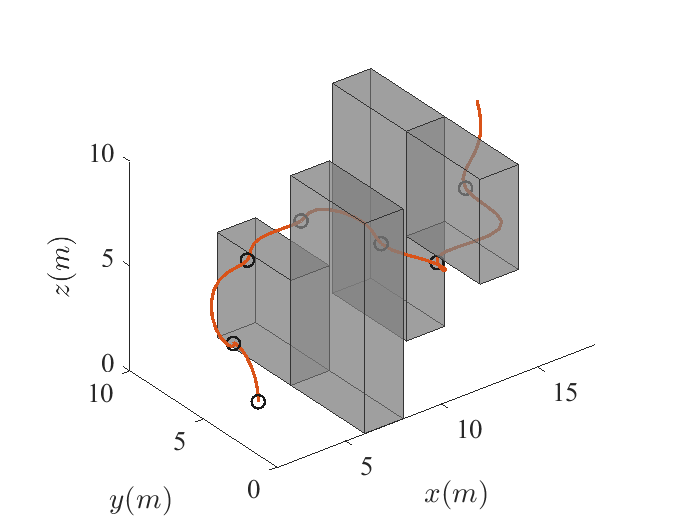}
     \end{subfigure}
        \caption{Kinodynamic RRT* results of the quadrotor. The first figure corresponds to the first solution found. 
        }
        \label{TreeQuadcomp1}
\end{figure}

\begin{figure}
    \centering
    \begin{subfigure}[b]{0.49\columnwidth}
         \centering
         \includegraphics[width=1\columnwidth]{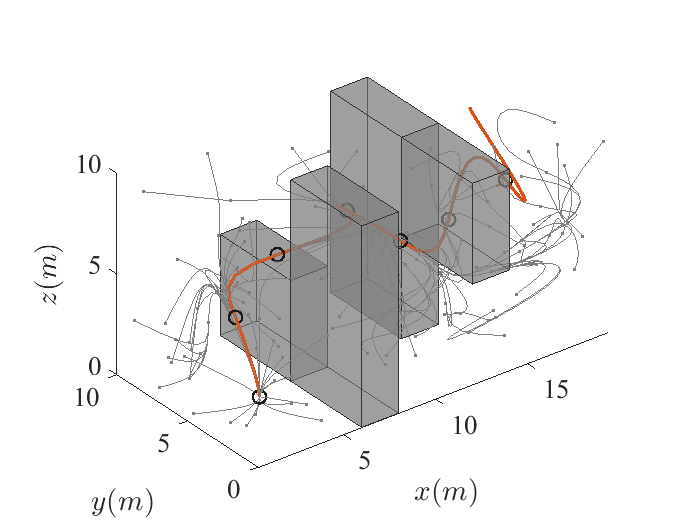}
     \end{subfigure}
     \begin{subfigure}[b]{0.49\columnwidth}
         \centering
         \includegraphics[width=1\columnwidth]{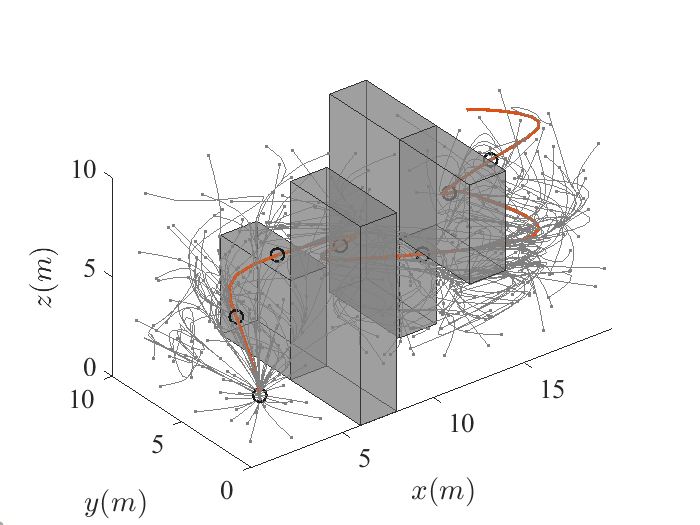}
     \end{subfigure}
     \begin{subfigure}[b]{0.49\columnwidth}
         \centering
         \includegraphics[width=1\columnwidth]{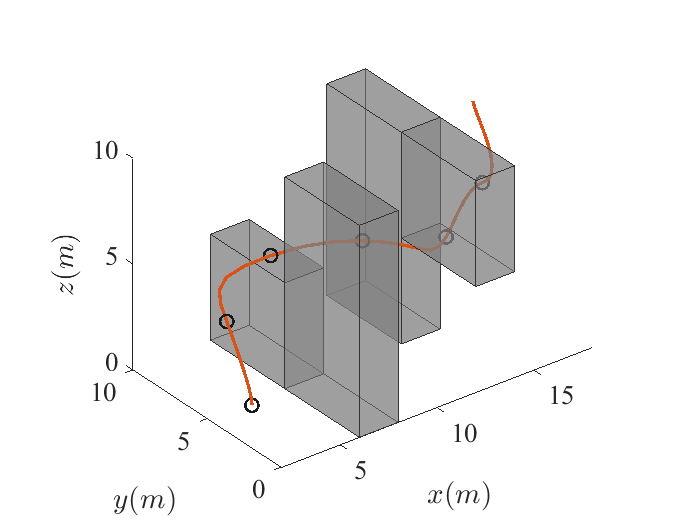}
     \end{subfigure}
     \begin{subfigure}[b]{0.49\columnwidth}
         \centering
         \includegraphics[width=1\columnwidth]{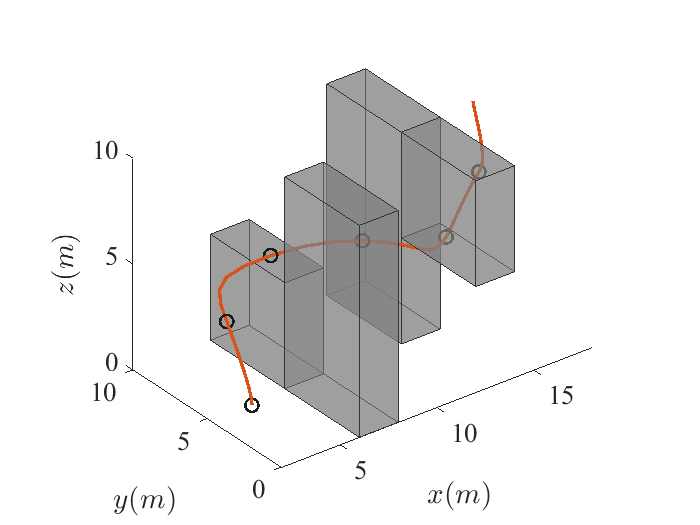}
     \end{subfigure}
        \caption{Kino-RRT* results of the quadrotor. 
        The first figure corresponds to the first solution found. 
        }
        \label{TreeQuadcomp2}
\end{figure}

\begin{figure}
    \centering
    \includegraphics[width=0.55\columnwidth]{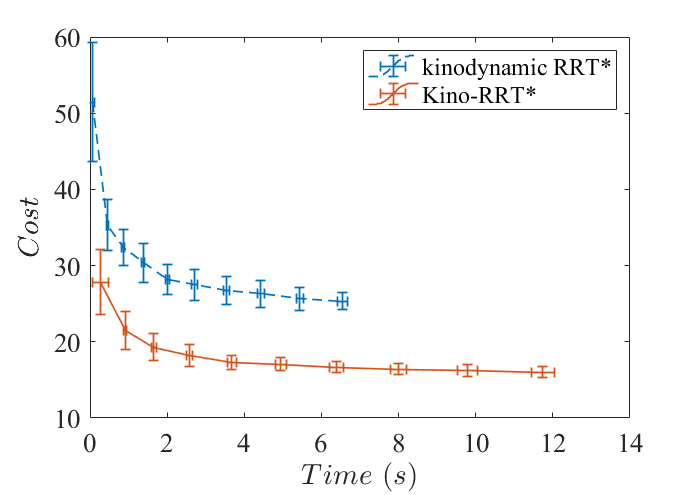}
        \caption{Comparison of Kino-RRT* and kinodynamic RRT* for the linearized quadrotor.}
        \label{CostQuadcomp}
\end{figure}

The results of the kinodynamic RRT* algorithm and the Kino-RRT* algorithm are given in Figure~\ref{TreeQuadcomp1} and Figure~\ref{TreeQuadcomp2}, respectively.
In Figure~\ref{TreeQuadcomp1}, from the upper left to bottom right, the number of nodes in the tree are $159$, $400$, $1000$, $2000$. 
The corresponding time to generate these trees are $0.147$, $0.48$, $1.92$, $6.18$ sec. 
The cost of the best trajectory in these trees are $58.10$, $30.92$, $24.84$, $24.61$, respectively.
In Figure~\ref{TreeQuadcomp1}, from upper left to bottom right, the number of nodes in the tree are $133$, $400$, $1000$, $2000$. 
The corresponding time to generate these trees are $0.19$, $0.84$, $3.72$, $11.35$ sec. 
The cost of the best trajectory in these trees are $20.31$, $19.13$, $15.56$, $15.52$, respectively.
The comparison of Kino-RRT* and kinodynamic RRT* is shown in Figure~\ref{CostQuadcomp}. 
The solution of the PPF controller with quadratic  terminal penalty is more complex than the fixed-final-state free-final-time controller. 
Thus, the Kino-RRT* algorithm takes more time to expand the same number of nodes compared to the kinodynamic RRT*.
Because each node in Kino-RRT* is more optimized, it still converges faster than the kinodynamic RRT*. 

\begin{figure}
    \centering
    \begin{subfigure}[b]{0.49\columnwidth}
         \centering
         \includegraphics[width=1\columnwidth]{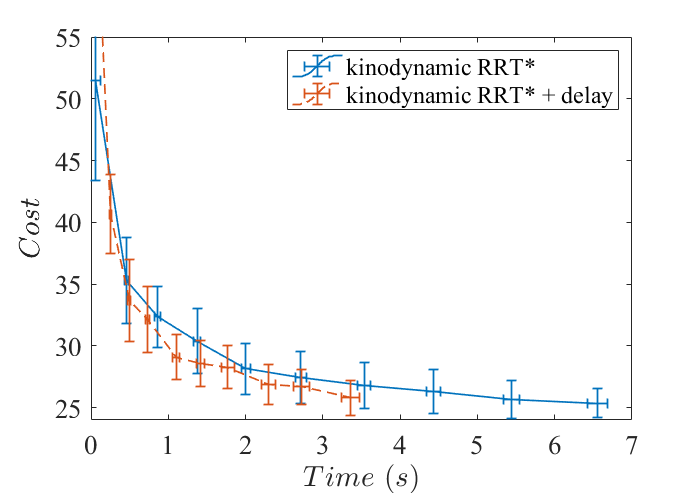}
     \end{subfigure}
     \begin{subfigure}[b]{0.49\columnwidth}
         \centering
         \includegraphics[width=1\columnwidth]{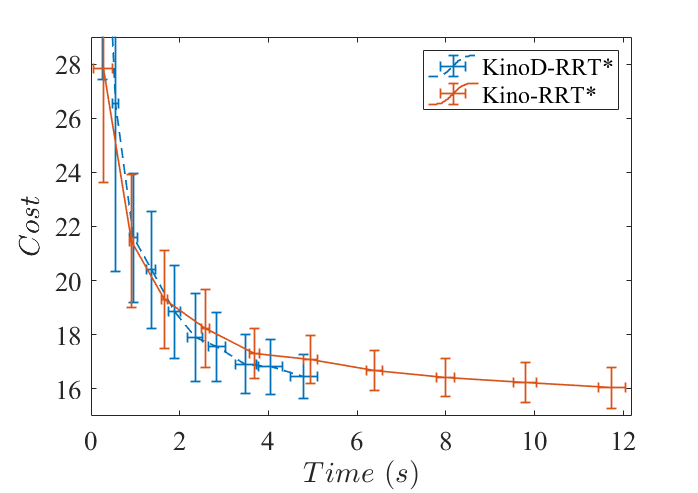}
     \end{subfigure}
        \caption{Delayed and intermittent update of the optimal arrival time for the linearized quadrotor example.}
        \label{DelayedQuadcomp}
\end{figure}

Figure~\ref{DelayedQuadcomp} shows the results of the delayed and intermittent update of the optimal arrival time. 
For the linearized quadrotor example, the kinodynamic RRT* with delay is 2 times faster than the kinodynamic RRT* when expanding the same number of nodes, and is also 2 times faster for finding a trajectory with a similar cost.
Similar performance improvement is observed for the KinoD-RRT* compared to Kino-RRT*. 
This performance improvement depends on the heuristic of the arrival time for the KinoD-RRT* algorithm.  

\section{Conclusion}  \label{secConclusion}

In this paper, we developed the Kino-RRT* algorithm, which utilizes a partial-final-state-free (PFF) optimal controller to improve the convergence performance of sampling-based motion planning of kinodynamic systems.
Instead of sampling the full state of the robot, Kino-RRT* only samples part of the state-space and the rest of the states are optimized by the PFF optimal controller. 
Although the algorithm is demonstrated on linear systems, the idea of PFF can be used as in~\cite{Xie2015Towards,Wolfslag2018RRT-CoLearn,zheng2021sampling}  for nonlinear kinodynamic systems as well.
We tested the algorithm on robot systems with 4-D and 10-D state-spaces.
In both cases, Kino-RRT* showed better convergence compared to the standard kinodynamic RRT*, achieving trajectories with better cost using much less time to compute. 
The proposed Kino-RRT* algorithm shows potential in real-time kinodynamic motion planning for high-dimensional dynamical systems.


\bibliographystyle{ieeetr}
\bibliography{references}

\begin{thebibliography}{10}

\bibitem{Dustin2013Kinodynamic}
D.~J. Webb and J.~Van Den~Berg, ``Kinodynamic {RRT*}: Asymptotically optimal
  motion planning for robots with linear dynamics,'' in {\em IEEE International
  Conference on Robotics and Automation}, (Karlsr\"{u}he, Germany),
  pp.~5054--5061, May 2013.

\bibitem{lavalle2011motion}
S.~M. LaValle, ``Motion planning: Wild frontiers,'' {\em IEEE Robotics
  Automation Magazine}, vol.~18, no.~2, pp.~108--118, 2011.

\bibitem{lavalle2001randomized}
S.~M. LaValle and J.~J. Kuffner~Jr, ``Randomized kinodynamic planning,'' {\em
  The International Journal of Robotics Research}, vol.~20, no.~5,
  pp.~378--400, 2001.

\bibitem{Karaman2011Sampling}
S.~Karaman and E.~Frazzoli, ``Sampling-based algorithms for optimal motion
  planning,'' {\em The International Journal of Robotics Research}, vol.~30,
  pp.~846--894, June 2011.

\bibitem{Karaman2011Anytime}
S.~Karaman, M.~R. Walter, A.~Perez, E.~Frazzoli, and S.~Teller, ``Anytime
  motion planning using the {RRT*},'' in {\em IEEE International Conference on
  Robotics and Automation}, (Shanghai, China), pp.~1478--1483, May 2011.

\bibitem{gonzalez2015review}
D.~Gonz{\'a}lez, J.~P{\'e}rez, V.~Milan{\'e}s, and F.~Nashashibi, ``A review of
  motion planning techniques for automated vehicles,'' {\em IEEE Transactions
  on Intelligent Transportation Systems}, vol.~17, no.~4, pp.~1135--1145, 2015.

\bibitem{gammell4asymptotically}
J.~D. Gammell and M.~P. Strub, ``Asymptotically optimal sampling-based motion
  planning methods,'' {\em Annual Review of Control, Robotics, and Autonomous
  Systems}, vol.~4, pp.~1--25, 2021.

\bibitem{Karaman2010Optimal}
S.~Karaman and E.~Frazzoli, ``Optimal kinodynamic motion planning using
  incremental sampling-based methods,'' in {\em 49th IEEE Conference on
  Decision and Control}, (Atlanta, GA), pp.~7681--7687, December 2010.

\bibitem{Gammell2014InformedRRT}
J.~D. Gammell, S.~S. Srinivasa, and T.~D. Barfoot, ``{Informed RRT*}: Optimal
  sampling-based path planning focused via direct sampling of an admissible
  ellipsoidal heuristic,'' in {\em IEEE/RSJ International Conference on
  Intelligent Robots and Systems}, (Chicago, IL), pp.~2997--3004, September
  2014.

\bibitem{Perez2012LQR}
A.~Perez, R.~Platt, G.~Konidaris, L.~Kaelbling, and T.~Lozano-Perez,
  ``{LQR-RRT*}: Optimal sampling-based motion planning with automatically
  derived extension heuristics,'' in {\em IEEE International Conference on
  Robotics and Automation}, (Saint Paul, MN), pp.~2537--2542, May 2012.

\bibitem{Wolfslag2018RRT-CoLearn}
W.~J. Wolfslag, M.~Bharatheesha, T.~M. Moerland, and M.~Wisse, ``{RRT-CoLearn}:
  Towards kinodynamic planning without numerical trajectory optimization,''
  {\em IEEE Robotics and Automation Letters}, vol.~3, no.~3, pp.~1655--1662,
  2018.

\bibitem{zheng2021sampling}
D.~Zheng and P.~Tsiotras, ``Sampling-based kinodynamic motion planning using a
  neural network controller,'' in {\em AIAA Scitech 2021 Forum}, p.~1754, 2021.

\bibitem{Chiang2019RLRRT}
H.~T.~L. Chiang, J.~Hsu, M.~Fiser, L.~Tapia, and A.~Faust, ``{RL-RRT}:
  Kinodynamic motion planning via learning reachability estimators from {RL}
  policies,'' {\em IEEE Robotics and Automation Letters}, vol.~4, no.~4,
  pp.~4298--4305, 2019.

\bibitem{arslan2015machine}
O.~Arslan and P.~Tsiotras, ``Machine learning guided exploration for
  sampling-based motion planning algorithms,'' in {\em IEEE/RSJ International
  Conference on Intelligent Robots and Systems}, (Hamburg, Germany),
  pp.~2646--2652, 2015.

\bibitem{janson2015fast}
L.~Janson, E.~Schmerling, A.~Clark, and M.~Pavone, ``Fast marching tree: A fast
  marching sampling-based method for optimal motion planning in many
  dimensions,'' {\em The International Journal of Robotics Research}, vol.~34,
  no.~7, pp.~883--921, 2015.

\bibitem{paden2017verification}
B.~Paden, V.~Varricchio, and E.~Frazzoli, ``Verification and synthesis of
  admissible heuristics for kinodynamic motion planning,'' {\em IEEE Robotics
  and Automation Letters}, vol.~2, no.~2, pp.~648--655, 2017.

\bibitem{yi2018generalizing}
D.~Yi, R.~Thakker, C.~Gulino, O.~Salzman, and S.~Srinivasa, ``Generalizing
  informed sampling for asymptotically-optimal sampling-based kinodynamic
  planning via markov chain monte carlo,'' in {\em IEEE International
  Conference on Robotics and Automation}, (Brisbane, Australia),
  pp.~7063--7070, 2018.

\bibitem{Xie2015Towards}
C.~Xie, J.~van~den Berg, S.~Patil, and P.~Abbeel, ``Toward asymptotically
  optimal motion planning for kinodynamic systems using a two-point boundary
  value problem solver,'' in {\em IEEE International Conference on Robotics and
  Automation}, (Seattle, WA), pp.~4187--4194, May 2015.

\bibitem{karaman2013sampling}
S.~Karaman and E.~Frazzoli, ``Sampling-based optimal motion planning for
  non-holonomic dynamical systems,'' in {\em 2013 IEEE International Conference
  on Robotics and Automation}, (Karlsr\"{u}he, Germany), pp.~5041--5047, 2013.

\bibitem{schmerling2015optimaldriftless}
E.~Schmerling, L.~Janson, and M.~Pavone, ``Optimal sampling-based motion
  planning under differential constraints: the driftless case,'' in {\em IEEE
  International Conference on Robotics and Automation}, (Seattle, WA),
  pp.~2368--2375, 2015.

\bibitem{schmerling2015optimaldrift}
E.~Schmerling, L.~Janson, and M.~Pavone, ``Optimal sampling-based motion
  planning under differential constraints: the drift case with linear affine
  dynamics,'' in {\em 54th IEEE Conference on Decision and Control}, (Osaka,
  Japan), pp.~2574--2581, 2015.

\bibitem{goretkin2013optimal}
G.~Goretkin, A.~Perez, R.~Platt, and G.~Konidaris, ``Optimal sampling-based
  planning for linear-quadratic kinodynamic systems,'' in {\em 2013 IEEE
  International Conference on Robotics and Automation}, (Karlsr\"{u}he,
  Germany), pp.~2429--2436, 2013.

\bibitem{sakcak2019sampling}
B.~Sakcak, L.~Bascetta, G.~Ferretti, and M.~Prandini, ``Sampling-based optimal
  kinodynamic planning with motion primitives,'' {\em Autonomous Robots},
  vol.~43, no.~7, pp.~1715--1732, 2019.

\bibitem{hauser2016asymptotically}
K.~Hauser and Y.~Zhou, ``Asymptotically optimal planning by feasible
  kinodynamic planning in a state--cost space,'' {\em IEEE Transactions on
  Robotics}, vol.~32, no.~6, pp.~1431--1443, 2016.

\bibitem{Li2016Asymptotically}
Y.~Li, Z.~Littlefield, and K.~E. Bekris, ``Asymptotically optimal
  sampling-based kinodynamic planning,'' {\em The International Journal of
  Robotics Research}, vol.~35, no.~5, pp.~528--564, 2016.

\bibitem{Sivaramakrishnan2019}
A.~Sivaramakrishnan, Z.~Littlefield, and K.~E. Bekris, ``Towards learning
  efficient maneuver sets for kinodynamic motion planning,'' {\em arXiv
  preprint, arXiv:1907.07876}, 2019.

\bibitem{li2021mpc}
L.~Li, Y.~Miao, A.~H. Qureshi, and M.~C. Yip, ``{MPC-MPNet}: Model-predictive
  motion planning networks for fast, near-optimal planning under kinodynamic
  constraints,'' {\em arXiv preprint arXiv:2101.06798}, 2021.

\bibitem{lewis2012optimal}
F.~L. Lewis, D.~Vrabie, and V.~L. Syrmos, {\em Optimal Control}.
\newblock John Wiley \& Sons, 2012.

\end{thebibliography}

\addtolength{\textheight}{-12cm}   

\end{document}